\documentclass[10pt,journal,compsoc]{IEEEtran}
\ifCLASSOPTIONcompsoc
  \usepackage[nocompress]{cite}
\else
  \usepackage{cite}
\fi
\ifCLASSINFOpdf
\else
\fi

\usepackage{color}
\usepackage{amsmath}
\usepackage{graphicx}
\usepackage{bm}
\usepackage{multirow}
\usepackage{makecell}
\usepackage{ragged2e}
\usepackage{balance}

\newcommand{\secref}[1]{Section~\ref{sec:#1}}
\newcommand{\figref}[1]{Figure~\ref{fig:#1}}
\newcommand{\tabref}[1]{Table~\ref{tab:#1}}
\newcommand{\eqnref}[1]{Eq.~\eqref{eq:#1}}

\newenvironment{myitemize2}[1][]{%%%%%¶¨ÒåÐÂ»·¾³
	
	\begin{list}{$\bullet$}
		{
			\setlength{\leftmargin}{5mm} %×ó±ß½ç
			\setlength{\parsep}{0.5mm} %¶ÎÂä¼ä¾à
			\setlength{\topsep}{0mm} %ÁÐ±íµ½ÉÏÏÂÎÄµÄ´¹Ö±¾àÀë
			\setlength{\itemsep}{0mm} %±êÇ©¼ä¾à
			\setlength{\labelsep}{0.5em} %±êºÅºÍÁÐ±íÏîÖ®¼äµÄ¾àÀë,Ä¬ÈÏ0.5em
			\setlength{\itemindent}{0mm} %±êÇ©Ëõ½øÁ¿
			\setlength{\listparindent}{6mm} %¶ÎÂäËõ½øÁ¿
	}}
	{\end{list}}%%%%%

\begin{document}
%
% paper title
% Titles are generally capitalized except for words such as a, an, and, as,
% at, but, by, for, in, nor, of, on, or, the, to and up, which are usually
% not capitalized unless they are the first or last word of the title.
% Linebreaks \\ can be used within to get better formatting as desired.
% Do not put math or special symbols in the title.
\title{Weakly-Supervised Video Object Grounding \\ via Causal Intervention}
%
%
% author names and IEEE memberships
% note positions of commas and nonbreaking spaces ( ~ ) LaTeX will not break
% a structure at a ~ so this keeps an author's name from being broken across
% two lines.
% use \thanks{} to gain access to the first footnote area
% a separate \thanks must be used for each paragraph as LaTeX2e's \thanks
% was not built to handle multiple paragraphs
%
%
%\IEEEcompsocitemizethanks is a special \thanks that produces the bulleted
% lists the Computer Society journals use for "first footnote" author
% affiliations. Use \IEEEcompsocthanksitem which works much like \item
% for each affiliation group. When not in compsoc mode,
% \IEEEcompsocitemizethanks becomes like \thanks and
% \IEEEcompsocthanksitem becomes a line break with idention. This
% facilitates dual compilation, although admittedly the differences in the
% desired content of \author between the different types of papers makes a
% one-size-fits-all approach a daunting prospect. For instance, compsoc 
% journal papers have the author affiliations above the "Manuscript
% received ..."  text while in non-compsoc journals this is reversed. Sigh.

\author{Wei~Wang,
        Junyu~Gao,
        and Changsheng~Xu,~\IEEEmembership{Fellow,~IEEE}% <-this % stops a space
\IEEEcompsocitemizethanks{\IEEEcompsocthanksitem Wei Wang and Junyu Gao are with National Lab of Pattern Recognition, Institute of Automation, Chinese Academy of Sciences, Beijing, 100190, P.R.China, and with School of Artifical Intelligence, University of Chinese Academy of Sciences, Beijing, China\protect\\
% note need leading \protect in front of \\ to get a newline within \thanks as
% \\ is fragile and will error, could use \hfil\break instead.
(E-mail: wangwei2018@ia.ac.cn; junyu.gao@nlpr.ia.ac.cn)
\IEEEcompsocthanksitem Changsheng Xu is with National Lab of Pattern Recognition, Institute of Automation, Chinese Academy of Sciences, Beijing, 100190, P.R.China, and with School of Artifical Intelligence, University of Chinese Academy of Sciences, Beijing, China, and also with Peng Cheng Laboratory, Shenzhen, China

(E-mail: csxu@nlpr.ia.ac.cn)}
\thanks{Manuscript received Oct 5, 2021.}}

% note the % following the last \IEEEmembership and also \thanks - 
% these prevent an unwanted space from occurring between the last author name
% and the end of the author line. i.e., if you had this:
% 
% \author{....lastname \thanks{...} \thanks{...} }
%                     ^------------^------------^----Do not want these spaces!
%
% a space would be appended to the last name and could cause every name on that
% line to be shifted left slightly. This is one of those "LaTeX things". For
% instance, "\textbf{A} \textbf{B}" will typeset as "A B" not "AB". To get
% "AB" then you have to do: "\textbf{A}\textbf{B}"
% \thanks is no different in this regard, so shield the last } of each \thanks
% that ends a line with a % and do not let a space in before the next \thanks.
% Spaces after \IEEEmembership other than the last one are OK (and needed) as
% you are supposed to have spaces between the names. For what it is worth,
% this is a minor point as most people would not even notice if the said evil
% space somehow managed to creep in.

% The paper headers
\markboth{Journal of \LaTeX\ Class Files,~Vol.~14, No.~8, August~2015}%
{Shell \MakeLowercase{\textit{et al.}}: Bare Demo of IEEEtran.cls for Computer Society Journals}
% The only time the second header will appear is for the odd numbered pages
% after the title page when using the twoside option.
% 
% *** Note that you probably will NOT want to include the author's ***
% *** name in the headers of peer review papers.                   ***
% You can use \ifCLASSOPTIONpeerreview for conditional compilation here if
% you desire.

% The publisher's ID mark at the bottom of the page is less important with
% Computer Society journal papers as those publications place the marks
% outside of the main text columns and, therefore, unlike regular IEEE
% journals, the available text space is not reduced by their presence.
% If you want to put a publisher's ID mark on the page you can do it like
% this:
%\IEEEpubid{0000--0000/00\$00.00~\copyright~2015 IEEE}
% or like this to get the Computer Society new two part style.
%\IEEEpubid{\makebox[\columnwidth]{\hfill 0000--0000/00/\$00.00~\copyright~2015 IEEE}%
%\hspace{\columnsep}\makebox[\columnwidth]{Published by the IEEE Computer Society\hfill}}
% Remember, if you use this you must call \IEEEpubidadjcol in the second
% column for its text to clear the IEEEpubid mark (Computer Society jorunal
% papers don't need this extra clearance.)

% use for special paper notices
%\IEEEspecialpapernotice{(Invited Paper)}

% for Computer Society papers, we must declare the abstract and index terms
% PRIOR to the title within the \IEEEtitleabstractindextext IEEEtran
% command as these need to go into the title area created by \maketitle.
% As a general rule, do not put math, special symbols or citations
% in the abstract or keywords.
\IEEEtitleabstractindextext{%
\begin{abstract}
	\justifying We target at the task of weakly-supervised video object grounding (WSVOG), where only video-sentence annotations are available during model learning. It aims to localize objects described in the sentence to visual regions in the video, which is a fundamental capability needed in pattern analysis and machine learning. Despite the recent progress, existing methods all suffer from the severe problem of spurious association, which will harm the grounding performance. In this paper, we start from the definition of WSVOG and pinpoint the spurious association from two aspects: (1) the association itself is not object-relevant but extremely ambiguous due to weak supervision, and (2) the association is unavoidably confounded by the observational bias when taking the statistics-based matching strategy in existing methods. With this in mind, we design a unified causal framework to learn the deconfounded object-relevant association for more accurate and robust video object grounding. Specifically, we learn the object-relevant association by causal intervention from the perspective of video data generation process. To overcome the problems of lacking fine-grained supervision in terms of intervention, we propose a novel spatial-temporal adversarial contrastive learning paradigm. To further remove the accompanying confounding effect within the object-relevant association, we pursue the true causality by conducting causal intervention via backdoor adjustment. Finally, the deconfounded object-relevant association is learned and optimized under a unified causal framework in an end-to-end manner. Extensive experiments on both IID and OOD testing sets of three benchmarks demonstrate its accurate and robust grounding performance against state-of-the-arts.

\end{abstract}

% Note that keywords are not normally used for peerreview papers.
}

% make the title area
\maketitle

% To allow for easy dual compilation without having to reenter the
% abstract/keywords data, the \IEEEtitleabstractindextext text will
% not be used in maketitle, but will appear (i.e., to be "transported")
% here as \IEEEdisplaynontitleabstractindextext when the compsoc 
% or transmag modes are not selected <OR> if conference mode is selected 
% - because all conference papers position the abstract like regular
% papers do.
\IEEEdisplaynontitleabstractindextext
% \IEEEdisplaynontitleabstractindextext has no effect when using
% compsoc or transmag under a non-conference mode.

% For peer review papers, you can put extra information on the cover
% page as needed:
% \ifCLASSOPTIONpeerreview
% \begin{center} \bfseries EDICS Category: 3-BBND \end{center}
% \fi
%
% For peerreview papers, this IEEEtran command inserts a page break and
% creates the second title. It will be ignored for other modes.
\IEEEpeerreviewmaketitle

\begin{figure}[tbp]
	\centering
	\includegraphics[width=\columnwidth]{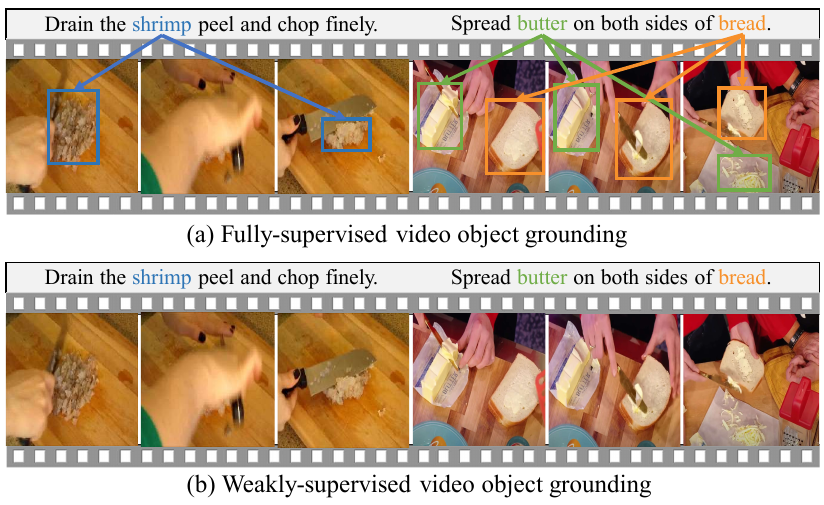}
	\vspace{-6mm}
	\caption{Two settings of video object grounding. They both aim to localize the objects described in the sentence to visual regions in a video. \mbox{(a) Fully-supervised} video object grounding is explicitly supervised with spatial bounding boxes and temporal object occurrences. (b) Weakly-supervised video object grounding only has video-sentence annotations during model learning, suffering from the ambiguity problem under weak supervision and confounding effect brought by co-occurred objects.} \label{fig:introduction}
	\vspace{-2mm}
\end{figure}

\IEEEraisesectionheading{\section{Introduction}\label{sec:introduction}}
	\IEEEPARstart{G}{rounding} natural language in visual regions has attracted increasing attention in recent years, because it is a fundamental capability needed for various downstream vision-and-language tasks, such as VQA~\cite{lei2019tvqa+}, robotics~\cite{alomari2017natural}, and image/video retrieval~\cite{karpathy2015deep}. It aims at establishing the instance-level correspondence between visual regions and textual objects described in sentences. For example, given an image/video and its natural language description ``spread butter on both sides of bread", the goal is to localize the described objects \emph{``butter''} and \emph{``bread''} to the corresponding visual regions. Recently, much progress has been made in context of static images, while visual grounding in video domain is still far from being fully explored. Compared with static images, videos are possessed of complex spatial-temporal structures, which impedes the direct application of those effective image grounding methods into video domain. Besides, collecting fine-grained region-level annotations for videos requires significant manual efforts and may bring in human errors. Therefore, as shown in~\figref{introduction}, we investigate the problem of video object grounding under weak supervision from only video-sentence annotations.

\vspace{1em}
Nevertheless, grounding a target object from such weak supervision is particularly challenging due to the lack of annotations in spatial bounding boxes and temporal object occurrences. To tackle this challenge, existing methods mainly resort to the frame-level Multiple Instance Learning (MIL) framework for weakly-supervised video object grounding (WSVOG) task, where each frame in spatial is treated as a bag of visual regions, and frames throughout the video are temporally aggregated with different weights so as to utilize the video-level supervision. Despite their promising results, these learning pipelines still suffer from severe matching ambiguity for object grounding in both spatial and temporal aspects, which greatly hinders the grounding performance. To be specific, in spatial, the popular \emph{maximum} MIL operator is adopted within each frame-bag to pick out the visual region that best matches the described object. However, as shown in~\figref{introduction} (b), due to the lack of spatial bounding box annotations, many distracting and background regions in the frame-bag will result in severe spatial ambiguity in inferring the positive object-relevant visual region. Similarly, in temporal, due to the lack of temporal object occurrence annotations, \textcolor{black}{the motion blur and shot cut in the video} will lead to severe temporal ambiguity in selecting the object-relevant frames for aggregation.

Unfortunately, both types of ambiguity are intrinsically possessed in the video for WSVOG task, which can never be completely removed from only video-sentence supervision, since there is no free lunch in the world. Although the conclusion seems disappointing, does it mean that we can do nothing but suffer? In this paper, we present the answer is “No”, by looking into such ambiguity from the perspective of video data generation process, and we contribute a formal principle to alleviate such ambiguity based on \textbf{causal analysis}~\cite{pearl2016causal}.  
We first formulate the WSVOG model as $P(Y|V,Q)$, which learns the association between the input video $V$ with description sentence $Q$ and the video-sentence label $Y$. The output of the model, which is the probability $p=P(Y|V,Q)$, is regarded as the video-level grounding score. By learning $P(Y|V,Q)$, the instance-level region-object correspondence is implicitly constructed, inducing a good grounding performance at test phase. During the weakly-supervised learning process, the two types of ambiguity mentioned before make it difficult to establish a strong object-relevant association in terms of regions and frames. In other words, for a description $Q$ containing a set of queried objects, many object-irrelevant regions and frames in video $V$ may contribute to a high grounding score, which shows the \emph{spurious} association between $V$ and label $Y$~\footnote{We will explain why $Q$ is not involved later in~\secref{causal_graph}.}. To further look into such spurious association, we investigate the data generation process of input video $V$ through a \textbf{causal} lens. Since there are a number of factors causing the generated video, we divide all the causal factors into two categories for simplicity. The object-relevant factors that cause the object's visual appearances in spatial and temporal throughout the video are grouped into a category, called \emph{content} $C$. The rest factors that cause, for example the background or scenes, are grouped into another category called \emph{style} $S$, which is independent of $C$. It can be formulated that $C\rightarrow V\leftarrow S$, where the directed links denote the causalities between two nodes. From a causal perspective, the principle of learning a stronger object-relevant association for the WSVOG model $P(Y|V,Q)$ is to undermine the causal effect of $S$ on $Y$, and retain the causal effect of $C$. We denote such improved association for the WSVOG model as $P_c(Y|V,Q)$.

Nevertheless, learning the object-relevant association $P_c(Y|V,Q)$ does not mean that we can safely establish the  region-object correspondence for WSVOG. Since WSVOG aims to ground multiple objects at a time, visual objects generated from \emph{content} $C$ in the video will co-occur with each other under certain contexts. Some of them are likely to co-occur more often (\emph{e.g.}, ``bread" and ``butter" in cooking videos), while some of them rarely co-occur (\emph{e.g.}, ``bread" and ``vinegar"). As a result, it is difficult to distinguish those frequently co-occurred objects in the video and thus correctly associate the textual “bread” object with the visual regions of “bread” rather than that of “butter”, since 
we only know there exist ``bread" and ``butter" from the description sentence but do not know where they are. Due to their frequent co-occurrences, 
the model will inevitably relate the visual regions of ``bread" with that of ``butter" when grounding. 
Ideally, if we could collect enough videos covering all the combinations of different object co-occurrences in a balanced distribution, we can distinguish any objects from them easily. However, it is labor-intensive or even impossible to collect such a huge dataset. 
On the other hand, we notice that the object co-occurrences essentially encode the natural dependencies of objects, since ``butter" is indeed a good meal companion for ``bread" rather than ``vinegar". This is helpful for a better association that it is likely a ``bread" region when seeing a ``butter" region. If we force ``bread" and ``vinegar" to frequently co-occur, it will harm such natural dependencies. Therefore, we have to respect the observed object co-occurrences in the dataset, and meanwhile remove the bad effect from it. 
To this end, we propose to use $P_c(Y|{do}(V,Q))$ instead of $P_c(Y|V,Q)$ to pursue the true causalities between the video-sentence label $Y$ and input video $V$ with its description $Q$, rather than the statistical associations of the observed data. The \emph{do}-operation denotes that we physically intervene the inputs to see what truly causes the label. The ideal way to calculate $P_c(Y|{do}(V,Q))$ is randomised controlled trial\cite{chalmers1981method}, by intervening the inputs to be under any possible context in videos\cite{dvornik2018modeling}. Thanks to the fact that we are learning the improved object-relevant association which has undermined the causal effect of \emph{style} $S$ on $Y$, we can represent the infinite possible contexts with finite objects in the dataset. With the above analysis, we can say that object in videos is essentially a \emph{confounder}~\cite{pearl2016causal} that misleads the grounding model to learn spurious association.

In light of the above observations, we propose a unified causal framework for WSVOG, which aims to learn the deconfounded object-relevant association $P_c(Y|{do}(V,Q))$ via causal intervention. We first take a causal view of WSVOG task, and formally pinpoint the exact causal paths that lead to the aforementioned ambiguity problem and confounding effect. Based on such causal analysis, we then learn the object-relevant association $P_c(Y|V,Q)$ by causal intervention from the perspective of video data generation process. To overcome the problems of lacking fine-grained supervision in terms of intervention, we propose a novel spatial-temporal adversarial contrastive learning paradigm, which jointly enjoys the merits of (a) learning from noisy intervened variables, (b) fully exploiting the interventional effect, and (c) well-adapted to spatial-temporal structures of videos. To further remove the accompanying bad confounding effect in $P_c(Y|V,Q)$, which is caused by co-occurred objects, we pursue the true causality $P_c(Y|{do}(V,Q))$ by conducting causal intervention via backdoor adjustment~\cite{pearl2016causal}. We tackle the challenge of collecting vague and diverse visual object as the confounder by learning its textual counterpart in cross-modal alignment as the substitute. Finally, the deconfounded object-relevant association $P_c(Y|{do}(V,Q))$ is learned and optimized under a unified causal framework in an end-to-end manner. Extensive experiments on both IID~\cite{vapnik1999overview} and OOD~\cite{teney2020value} testing sets demonstrate that our approach not only achieves better grounding performance but also has stronger \mbox{generalizability against distribution changes.}

Our main contributions can be summarized as follows:
\vspace{0.3em}

\begin{myitemize2}
	\item We propose a unified causal framework for WSVOG, in which the ambiguity problem brought by weak supervision as well as the accompanying spurious association caused by confounder are clearly revealed and elegantly addressed. By conducting causal intervention, a deconfounded object-relevant association is learned \mbox{in an end-to-end manner.} 
	\item To the best of our knowledge, in the field of causal analysis in computer vision tasks, we are among the first to explicitly take the ambiguity problem brought by weak supervision into consideration, rather than simply attributing all the spurious association to a certain confounder. Accordingly, we propose a novel spatial-temporal adversarial contrastive learning paradigm, which can seamlessly cooperate with the subsequent deconfounding operation under a unified causal framework. Through theoretical analysis and extensive experiments, both of them are indispensable in WSVOG.
	\item We conduct extensive experiments on the IID and OOD testing sets of three benchmarks for WSVOG. Experimental results demonstrate that our method is capable of performing more accurate and robust video object grounding than state-of-the-arts.
\end{myitemize2}

\section{Related Work}
\subsection{Visual Grounding}
Typically, it aims at learning the correspondence between visual regions and textual entities in static images. In recent years, with the development of deep learning as well as the collection of large-scale visual grounding datasets, this task has achieved remarkable success. In particular, based on the densely annotated datasets like Flickr30K Entities~\cite{plummer2015flickr30k}, ReferItGame~\cite{kazemzadeh2014referitgame}, and Visual Genome~\cite{krishna2017visual}, many approaches learn a similarity function between regions and textual entities by using the ground-truth bounding box supervision~\cite{wang2018learning,chen2017msrc,plummer2017phrase,sadhu2020video,wang2016structured,sun2021discriminative,hong2019learning0,plummer2020revisiting}. Although these methods show promising performance on visual grounding, they highly rely on the dense annotation, which is too expensive to obtain in reality. 
Hence, weakly-supervised visual grounding is now becoming the research focus~\cite{chen2019object,chen2018knowledge,xiao2017weakly,zhao2018weakly,datta2019align2ground,liu2019knowledge}, since it is under a more general setting where only image-sentence annotations are needed for training.  Karpathy and Fei-Fei~\cite{karpathy2015deep} tackle this problem by calculating pairwise region-phrase similarities to rank the region proposals under multiple instance learning (MIL) framework. GroundR~\cite{rohrbach2016grounding} builds region-phrase correspondences by reconstructing the phrases from predicted region proposals, so as to utilize the attention scores of regions for grounding. Similarly, KAC-Net~\cite{chen2018knowledge} leverages visual consistency and external complementary knowledge from object categories to obtain more supervision. Follow the line of exploring more supervisory signals, Xiao \emph{et al.}~\cite{xiao2017weakly} use parse tree structures of sentences to supervise the region and phrase encoding consistency. Align2Ground~\cite{datta2019align2ground} utilizes the image caption task to obtain stronger supervisory signals. Liu \emph{et al.}~\cite{liu2021relation} exploit the supervision of context cues of visual object relations, so as to refine the region proposals. Recently, employing the popular contrastive learning strategy to better distinguish positive and negative pairs is another research line~\cite{gupta2020contrastive,zhang2020counterfactual,wang2021improving}. 
Although much progress has been made in context of static images, only limited attention has been paid in video domain. Due to the complex spatial-temporal structures of videos, these successful visual grounding \mbox{methods cannot be directly applied to video domain.}

\subsection{Weakly-Supervised Video Object Grounding}
There are several different settings in existing works in terms of video object grounding, including localizing the queried objects described in the sentence to visual regions in the video~\cite{huang2018finding,yu2017sentence,zhou2018weakly,shi2019not,yang2020weakly}, identifying spatial-temporal tubes within videos from referring expressions~\cite{zhang2020object,zhang2020does}, and localizing only the specific objects referred in the query sentence to visual regions~\cite{sadhu2020video}. Note that the last two settings are usually under full supervision. Due to the expensive cost of collecting fine-grained annotations in videos, this paper mainly targets at the first setting in a weakly-supervised fashion, where the objects described in the sentence are grounded to visual regions in the video from only video-sentence supervision.
To tackle this task, some earliest attempts are either restricted to constraint videos or simply extend the image-domain grounding methods to videos, which are unpractical and sub-optimal. Specifically, Yu \emph{et al.}~\cite{yu2017sentence} try to ground objects in constraint videos that are recorded in the laboratory. Huang \emph{et al.}~\cite{huang2018finding} propose to extend the image-domain MIL grounding framework DVSA~\cite{karpathy2015deep} to video domain for grounding the linguistic references in videos. It was not until Zhou \emph{et al.}~\cite{zhou2018weakly} that for the first time formally formulate this task as a frame-level MIL problem and collect a standard unconstraint dataset YouCook2-BB for it. Based on their work, Shi \emph{et al.}~\cite{shi2019not} take a step further by complementing the evaluation metrics of this task, and they build a strong baseline model by exploiting the visual contexts of each frame in temporal. Furthermore, STVG~\cite{yang2020weakly} improves the model from the probabilistic perspective, by simultaneously considering the spatial-temporal contextual similarities within the video in an end-to-end manner. Besides, Chen \emph{et al.}\cite{chen2020activity} choose to utilize external supervisory signals to boost the performance, where they exploit visual and textual activity cues under the supervision of external human pose detectors and powerful language models. Although these methods achieve promising results, they all focus on designing more delicate model structures, but take the intrinsic ambiguity brought by weak supervision as granted. Besides, they also neglect the long-held confounding effect within this task, and they are all learning the confounded associations in fact. In our unified causal framework, the ambiguity problem brought by weak supervision and confounded associations caused by the confounder are both elegantly addressed.

\subsection{Causal Analysis}
Causal analysis has been widely adopted in economics, psychology, and epidemiology for years~\cite{pearl2018book,brown1994economic,mackinnon2007mediation,richiardi2013mediation}, and it is now attracting increasing attention in the computer vision society. Causal analysis not only serves as an interpretation framework which reveals the intrinsical causalities within the model, but also provides guidance to remove the spurious association and thus disentangles the desired model effect by pursuing the true causal effect. Recently, it has been successfully applied in many computer vision tasks in terms of static images, such as image captioning~\cite{yang2020deconfounded}, image segmentation~\cite{zhang2020causal}, few-shot image classification~\cite{yue2020interventional}, long-tailed image classification~\cite{tang2020long}, visual dialogue~\cite{qi2020two}, region-based CNN~\cite{wang2020visual}, self-supervised representation learning~\cite{mitrovic2020representation,von2021self}, adversarial representation learning~\cite{zhang2021adversarial,tang2021adversarial}, and so on. While the image domain is well-researched, there are few works in context of spatial-temporally complex videos, where only video moment retrieval~\cite{yang2021deconfounded,nan2021interventional}, video action localization~\cite{liu2021blessings}, etc. are explored. Specifically, for video moment retrieval, Yang \emph{et al.}~\cite{yang2021deconfounded} propose that the confounder is temporal location of moments within the video, and they remove such confounding effect by causal intervention. Nan \emph{et al.}~\cite{nan2021interventional} think the spurious association mainly comes from the selection bias of the dataset that cannot be observed, and they build a surrogate confounder set from the vocabulary of captions so as to adopt causal intervention. However, none of them explicitly look into such confounding effect in aspect of complex spatial-temporal \mbox{structures of videos, which are unique in video domain.}

Nonetheless, from another perspective of supervision signals, most of the causal analysis works are under fully-supervised setting, while only few works explore the weakly-supervised setting. It is more challenging to learn the true causal effect under weak supervision, because the learned association is not just simply confounded by a certain confounder, but the association itself is extremely ambiguous. As far as we know, there are only two related causal analysis works under weak supervision, namely weakly-supervised semantic segmentation (WSSS)~\cite{zhang2020causal} and weakly-supervised action localization (WTAL)~\cite{liu2021blessings}. For WSSS, Zhang \emph{et al.}~\cite{zhang2020causal} attribute the vague boundary of objects' pseudo-masks in CAM to the confounding effect brought by context prior, but they neglect the cases where the pseudo-masks generated by CAM may not even correctly cover the regions of object due to the ambiguity brought by weak supervision. For WTAL, Liu \emph{et al.}~\cite{liu2021blessings} point out that the ``background" issue is impossible to be fundamentally resolved due to weak supervision, but they simply attribute every spurious association to an unobserved confounder and propose to learn a substitute by adopting PCA on videos. However, the derivation of utilizing PCA for deconfounding is based on many strict assumptions, which is not so practical in reality. Different from existing methods, we propose a unified causal framework which simultaneously addresses the ambiguity problem brought by weak supervision and the accompanying spurious association \mbox{caused by confounder without the loss of generality.}
\vspace{-1em}

\section{Causal View of WSVOG}\label{sec:causal_view}
In this section, we first model the causalities within WSVOG task by constructing a unified causal graph~\cite{pearl2000models}, where the intrinsically possessed ambiguity in videos under weak supervision, as well as the spurious association brought by the confounder, are clearly revealed. Then, we use causal analysis to work out the fundamental solutions to them.

\vspace{-1em}
\subsection{Causal Graph} \label{sec:causal_graph}
\begin{figure}[tbp]
	\centering
	\includegraphics[width=0.7\columnwidth]{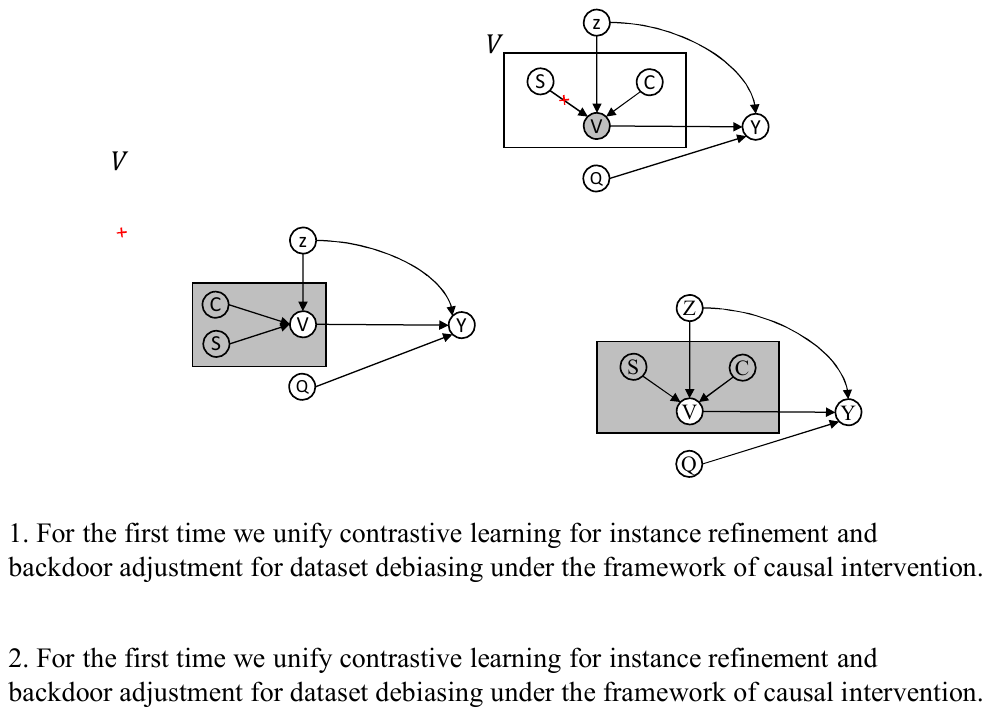}
	\vspace{-4mm}
	\caption{Our proposed unified causal graph of WSVOG task. The gray solid box models the data generation process of video $V$, in which the \emph{style} $S$ and \emph{content} $C$ are unobservable.} \label{fig:causal_graph}
	\vspace{-2mm}
\end{figure}

As illustrated in~\figref{causal_graph}, we construct a unified causal graph to model the causalities among four variables in WSVOG: video $V$, sentence $Q$, video-sentence label $Y$, and context (object) $Z$. It also models the causalities within the unobservable data generation process of video $V$ (shown in the gray solid box), where both \emph{style} $S$ and \emph{content} $C$ are unobservable variables. Note that we can also take this gray solid box as a whole and view it as $V$.  The causal graph is a directed acyclic graph where the directed link denotes the causality between two nodes. Conventional WSVOG methods only learn from two links: $V\rightarrow Y$ and $Q\rightarrow Y$, which ignore the ambiguity under weak supervision and spurious association brought by the confounder, while our method looks into these problems in a causal view and proposes the fundamental solutions to them.

\vspace{1em}
\noindent \bm{$S\rightarrow V\leftarrow C.$} This is the data generation process of video $V$ shown in the gray solid box. There are a number of factors causing the generated video $V$, which can be divided into \emph{content} $C$ that causes the objects' spatial-temporal visual appearances throughout the video, and \emph{style} $S$ that causes the background or scenes in the video. Both $S$ and $C$ are unobservable variables.

\vspace{1em}
\noindent \bm{$S\rightarrow V\rightarrow Y.$} The intrinsically possessed ambiguity in video $V$ for WSVOG task is mainly derived from this path. Due to the weak supervision, many object-irrelevant regions and frames generated from \emph{style} $S$ in video $V$ are involved in learning the association with video-sentence label $Y$, which are supposed to be eliminated, \emph{i.e.,} $S\not\rightarrow V\rightarrow Y$.

\vspace{1em}
\noindent \bm{$C\rightarrow V\rightarrow Y.$} We would like to learn a strong object-relevant association $P_c(Y|V,Q)$ where in video $V$ only object-relevant regions and frames generated from \emph{content} $C$ are supposed to be involved in. While such kind of association is easy to learn in fully-supervised setting, it can be extremely challenging under weak supervision. Therefore, our goal is to strengthen this path in WSVOG task.

\vspace{1em}
\noindent \bm{$V\leftarrow Z\rightarrow Y.$} 
This is a backdoor path~\cite{pearl2016causal} where the confounder $Z$ simultaneously affects the video $V$ and label $Y$, leading the model to learn the spurious association.
As discussed before, if we had successfully cut off the path $S\not\rightarrow V\rightarrow Y$, the learned object-relevant association $P_c(Y|V,Q)$ would only be confounded by co-occurred objects, because the potential confounder $Z$ is degraded from infinite possible contexts (from \emph{content} $C$ and \emph{style} $S$) to finite objects in the dataset (from \emph{content} $C$ only). The counfounder $Z$ eventually misleads the model via this backdoor path, where $Z\rightarrow V$ indicates that objects affect the representation learning of video $V$, and $Z\rightarrow Y$ indicates that co-occurred objects will affect the video-sentence label $Y$ (\emph{e.g.,} it is less likely a real video-sentence pair when objects ``bread" and ``vinegar" co-occur).

\vspace{1em}
\noindent \textbf{Why do the aforementioned problems not affect sentence $\bm Q$?} Because in WSVOG task, sentence $Q$ only provides what objects to be grounded in video $V$, which also can be seen as the \emph{query}. It is obvious that the aforementioned spatial and temporal ambiguity under weak supervision as well as the confounded association will only exist in \mbox{video $V$ rather than \emph{query} $Q$.}

\vspace{1em}
So far, we have pinpointed the exact causal paths that lead to the ambiguity in videos under weak supervision $(S\rightarrow V\rightarrow Y)$, and spurious association brought by the confounder $(V\leftarrow Z\rightarrow Y)$. Next, we will elaborate on how to cut off these paths using the power of causal intervention.

\subsection{Learning Object-Relevant Association}

To learn an improved object-relevant association $P_c(Y|V,Q)$ in place of the conventional association $P(Y|V,Q)$ that suffers from spatial and temporal ambiguity in WSVOG, we would like to cut off the path from \emph{style} $S$ to video $V$ ($S\not\rightarrow V\rightarrow Y$), and retain the path of $C\rightarrow V\rightarrow Y$.
Ideally, in such a model, the predicted probability $p_c=P_c(Y|V,Q)$, which is the grounding score, should be invariant to \emph{style} $S$ and sensitive to \emph{content} $C$. To formally formulate such properties in $P_c(Y|V,Q)$, we first introduce the concept of Interventional Effect (IE) that measures the effect on the output probability $p_c$ when intervening the generative factor ($S$ or $C$) to take a certain value ($\hat{s}$ or $\hat{c}$). Taking the generative factor \emph{style} $S$ as an example, the Interventional Effect is defined as:
\begin{equation}
	\begin{aligned}
		IE(p_c|do(S=\hat{s}))=P_c(Y|V,Q)-P_c^{do(S=\hat{s})}(Y|V,Q)\\
	\end{aligned} \label{eq:hsv}
\end{equation}
where the superscript \emph{do}-operation denotes that we do not directly intervene the inputs to take the certain value, but intervene the generative factors of them.

\vspace{1em}
\noindent Then we can easily obtain the following properties hold for every possible $\hat{s}$ and most of $\hat{c}$~\footnote{At most time, $\hat{c}$ is worse than its original value, thus leading to a decrease in grounding score $p_c$.}:
\begin{equation}
	\begin{aligned}
		&IE(p_c|do(S=\hat{s}))=0\\
		&IE(p_c|do(C=\hat{c}))>0\\
	\end{aligned} \label{eq:property}
\end{equation}
However, in practical, we are not able to directly or precisely manipulate the values of $S$ and $C$, because both of them are unobservable and there is no fine-grained supervision for us to distinguish them apart. To this end, we resort to the Contrastive Learning (CL) strategy as an alternative, which allows us to learn from noisy generative factors. Therefore, we reformulate~\eqnref{property} as the following inequality constraint:
\begin{equation}
	\begin{aligned}
		IE(p_c|do(S=\hat{s}))<IE(p_c|do(C=\hat{c}))\\
	\end{aligned} \label{eq:ineqa}
\end{equation}
By forcing the conventional association $P(Y|V,Q)$ to satisfy the above inequality constraint, we are able to estimate the object-relevant association $P_c(Y|V,Q)$.

In terms of implementation, we are still faced with some key problems that 1) how to effectively find out different types of generative factors, 2) how to set the proper values ($\hat{s}$ and $\hat{c}$) for these collected factors when performing the above inequality constraint, and 3) how to design a good Contrastive Learning framework to fully exploit the potential of the inequality constraint. We will elaborate our delicate implementation in~\secref{implementation_object_relevant}.

\subsection{\hspace{-2mm}\mbox{Removing Bad Effect in Object-Relevant Association}}
After we have cut off the path $S\not\rightarrow V\rightarrow Y$, the learned object-relevant association $P_c(Y|V,Q)$ is still confounded by co-occurred objects. As we have discussed before, object is essentially a confounder $Z$ that misleads the model to learn spurious association via the backdoor path $V\leftarrow Z\rightarrow Y$. Therefore, we propose to use $P_c(Y|do(V,Q))$ to pursue the true causality and remove the bad effect of confounder $Z$. Since the ideal way to calculate $P_c(Y|do(V,Q))$ --- intervening the inputs to be under any possible context --- is impossible, we apply the backdoor adjustment~\cite{pearl2016causal} to approximate it as follows:
\begin{equation}
	\begin{aligned}
		P_c(Y|do(V,Q))=\sum_{z\in Z}^{}P_c(Y|V,Q,z)P(z)\\
	\end{aligned} \label{eq:backdoor}
\end{equation}
The key idea behind it is to cut off the link $Z\rightarrow V$, and stratify the confounder $Z$ into pieces $Z=\{z\}$, so as to force the inputs to fairly interact with all possible cases of the confounder $Z$. From another perspective, backdoor adjustment brings in any possible object $z$ from other videos in the dataset, and then puts it into the current inputs so as to see what truly causes the label.

Nevertheless, although we are learning the object-relevant association, visual regions of the co-occurred objects in the video are still vague (\emph{e.g.,} ``bread" and ``butter"), which hinders us from cropping out the correct visual regions of other objects in the dataset for backdoor adjustment. We will present our implementation in~\secref{implementation_deconfound} to show how to overcome this problem.

\section{Methodology}
We first introduce the formulation of WSVOG task, then we present the detailed implementations of learning object-relevant association and removing bad effect in object-relevant association, respectively.

\subsection{Problem Formulation}\label{sec:MIL}
This task aims at localizing the described objects in the sentence to visual regions in the video with only video-sentence supervision. Formally, given a video $V$ consisting of $T$ frames, the $t_{\text{th}}$ frame contains a set of $N$ region proposals $R_t=\{r_n^t\}_{n=1}^N$. For its natural language description $Q$, it consists of $K$ described objects (in the form of phrases) $O=\{o_k\}_{k=1}^K$, where each object is supposed to be grounded to the corresponding visual regions in the video.

WSVOG task is mainly formulated as a frame-level MIL problem, where each frame in the video is regarded as a spatial frame-bag containing a set of region proposals, and frame-bags throughout the video are temporally aggregated with different weights. In particular, these two procedures correspond to Spatial Grounding and Temporal Localization, respectively. For Spatial Grounding, the matching similarities $M^k_{t,n}$ between visual regions and textual objects are first measured within each frame-bag. Then the popular \emph{maximum} MIL operator is adopted to obtain the frame-bag spatial grounding score $F^k_t$ as follows:
\begin{equation}
	\begin{aligned}
		M^k_{t,n}= \text{Sim}(o_k,r_n^t)\\
	\end{aligned}\label{eq:mtk}
\end{equation}
\begin{equation}
	\begin{aligned}
		F^k_t=\max \limits_{1\leq n \leq N} M^k_{t,n}\\
	\end{aligned}\label{eq:2}
\end{equation}
where $\text{Sim}(\cdot)$ is a similarity function.
In terms of Temporal Localization, each frame-bag in the video is assigned with an aggregation weight $Q_t^k$, which is calculated as follows:
\begin{equation}
	\begin{aligned}
		G^k_t=\text{Temp}( \{R_t\}_{t=1}^T,O)\\
	\end{aligned}\label{eq:temploc}
\end{equation}
where $\text{Temp}(\cdot)$ is a function that comprehensively considers the temporal region proposals and described objects, and produces the likelihood of each frame-bag being positive. Existing works implement this function in multiple ways~\cite{yang2020weakly,shi2019not,chen2020activity}, such as firstly concatenating the max-pooled features of region proposals and described objects and then utilizing MLPs with softmax to transform them into likelihood~\cite{yang2020weakly}.

Finally, the video-level grounding score $p=P(Y|V,Q)$ is the temporal aggregation of frame-bag grounding scores, followed with an average over all described objects:
\begin{equation}
	\begin{aligned}
		p=P(Y|V,Q)=\frac{1}{K} \sum_{k=1}^{K}  \sum_{t=1}^{T} F_t^k G_t^k\\
	\end{aligned} \label{eq:frameMIL}
\end{equation}
where $p$ can be directly supervised with the video-sentence label $Y$. In practice, this kind of frame-level MIL-based paradigm usually adopts ranking loss for a better learning, which is formulated as follows:
\begin{equation}
	\begin{aligned}
		\mathcal{L}_{\text{MIL}}=\ell(p_{[V'Q]}-p_{[VQ]})+\ell(p_{[VQ']}-p_{[VQ]})\\
	\end{aligned} \label{eq:newframeMIL}
\end{equation}
where $\ell(x)=\text{log}(1+\text{exp}(x/\eta))$, referring to the log loss with a scale parameter $\eta$. The subscript $[VQ]$ indicates the condition variables in $P(Y|V,Q)$. $V'$ and $Q'$ refer to the negative pair in the dataset.

\subsection{Implementation of Learning Object-Relevant Association} \label{sec:implementation_object_relevant}
According to the causal analysis stated before, we present the implementation of learning object-relevant association based on Contrastive Learning using the derived inequality constraint~\eqnref{ineqa}. 
We will elaborate our framework by sequentially answering the three previously raised problems.

\vspace{1em}
\noindent\textbf{Q1:} How to effectively find out different types of generative factors ($S$ and $C$)?

Since $S$ and $C$ are unobservable variables within the video data generation process, we propose to employ the observable regions and frames generated from $S$ and $C$ as the proxy of them, named proxy variables. To effectively find out the proxy variables of \emph{content} $C$, we design a selection strategy that for a described object $o_k$, we select top-$B_{r}$ regions in each frame-bag according to the matching similarities $M_{t,n}^k$ in Spatial Grounding (\eqnref{mtk}). Likewise, we select top-$B_{f}$ frames throughout the video according to the temporal aggregation weights $G_t^k$ in Temporal Localization (\eqnref{temploc}). The selected regions and frames are grouped into the region-level content set $\bm{U_t^c}=\{r_n^t\}_{n=1}^{B_r}$ and frame-level content set $\bm{H^c}=\{R_t\}_{t=1}^{B_f}$, respectively. The rest regions in each frame-bag, as well as the rest frames throughout the video, are grouped into the region-level style set $\bm{U_t^s}$ and frame-level style set $\bm{H^s}$, respectively. Here, we have to point out that the selected content sets and style sets are unavoidably noisy, because there is no fine-grained supervision for us to exactly tell them apart. Thanks to Contrastive Learning, we can still learn from such noisy data by using the derived inequality constraint~\eqnref{ineqa}.

\vspace{1em}
\noindent\textbf{Q2:} How to set the proper values for these collected proxy variables when performing the \emph{do}-operation in inequality constraint?

A simple way is to let the proxy variables of $S$ and $C$ be zeros (also can be seen as wiping them out), \emph{i.e.,} $do(\bm{H^c}=\bm{0})$ and $do(\bm{H^s}=\bm{0})$. However, this kind of intervention does not make full use of the inequality constraint in~\eqnref{ineqa}, because wiping out the top-ranked regions or frames will of course have a larger effect on grounding score than wiping out the bottom-ranked ones, which makes the inequality constraint easy to satisfy. Therefore, we propose to perform the \emph{do}-operation by generating the adversarial counterpart for the intervened proxy variable so as to challenge this constraint to improve model's discriminative ability. Specifically, we generate adversarial samples based on the memory bank via memory vector replacement. Taking the region-level content set $\bm{U_t^c}$ as an example, we first construct a region-level memory bank consisting of $J$ memory vectors $\{m_j\}_{j=1}^J$ which are initialized with randomly selected visual regions in the training set. Then, for each proxy variable $r_n^t$ in $\bm{U_t^c}$, we measure the similarities between $r_n^t$ and all the memory vectors $m_j$, and pick out its most similar memory vector as the adversarial sample $\hat{r}_n^t$:
\begin{equation}
	\begin{aligned}
		\hat{r}_n^t=\underset{m_j}{\text{argmax}}~\text{Sim}(r_n^t,m_j)\\
	\end{aligned} \label{eq:replacement}
\end{equation}
We denote such intervened set as $\bm{\hat{U}_t^c}$. Likewise, the intervened set of the rest region-level style set $\bm{U_t^s}$ in the $t_{th}$ frame-bag is denoted as $\bm{\hat{U}_t^s}$. 

For frame-level content set $\bm{H^c}$ and style set $\bm{H^s}$, we perform the \emph{do}-operation in a similar way. Note that we also construct a corresponding frame-level memory bank consisting of $W$ memory vectors for them and we measure the similarities in~\eqnref{replacement} based on the max-pooling of $R_t$ along the dimension of $N$. The intervened sets are denoted as $\bm{\hat{H}^c}$ and $\bm{\hat{H}^s}$, respectively.

\vspace{1em}
\noindent\textbf{Q3:} How to design a good Contrastive Learning framework to fully exploit the potential of the inequality constraint?

We design a spatial-temporal adversarial Contrastive Learning framework where the inequality constraint is fully applied and challenged in both region and frame levels throughout the video as follows:
\begin{equation}
	\begin{aligned}
		IE(p|do(\bm{U_t^s}=\bm{\hat{U}_t^s}))&<IE(p|do(\bm{U_t^c}=\bm{\hat{U}_t^c}))\\
		IE(p|do(\bm{H^s}=\bm{\hat{H}^s}))&<IE(p|do(\bm{H^c}=\bm{\hat{H}^c}))\\
	\end{aligned} \label{eq:contrast}
\end{equation}

To generate better adversarial samples that challenge the inequality constraint, we further update the memory banks during model learning. Taking the region-level memory bank $\{m_j\}_{j=1}^J$ as an example, we first calculate the average of all the proxy variables which have been replaced by $m_j$ in adversarial sample generation:
\begin{equation}
	\begin{aligned}
		a_j=\text{mean}(r_n^t~|~\hat{r}_n^t=m_j)\\
	\end{aligned} \label{eq:hsv}
\end{equation}
then we update the memory vectors in a momentum fashion:
\begin{equation}
	\begin{aligned}
		m_j\leftarrow\alpha m_j+(1-\alpha)a_j\\
	\end{aligned} \label{eq:hsv}
\end{equation}
where $\alpha$ is the momentum factor.

By applying the spatial-temporal adversarial contrastive learning, the memory banks are encouraged to help generate hard-to-distinguish adversarial samples to spatial-temporally challenge the inequality constraint, and the learned association is encouraged to satisfy its object-relevant Interventional Effect given these adversarial samples. After fully trained, we can learn the object-relevant association $P_c(Y|V,Q)$ in place of $P(Y|V,Q)$ for WSVOG.

\subsection{Implementation of Removing Bad Effect in Object-Relevant Association} \label{sec:implementation_deconfound}
As we have analyzed before, the learned object-relevant association $P_c(Y|V,Q)$ is still confounded by co-occurred objects in videos, and we aim to eliminate their confounding effect by adopting the backdoor adjustment in~\eqnref{backdoor}. To bring in every possible object $z$ from other videos in the dataset, we are faced with the problem that the visual regions of co-occurred objects are vague and thus they can hardly be correctly cropped out.

Fortunately, WSVOG itself is a cross-modal matching task where either vague or diverse visual regions of the same object are supposed to be aligned with its unified textual counterpart in common embedding space. In other words, the textual embedding of $o_k$ (usually the word embedding like GloVe~\cite{pennington2014glove}) essentially provides the stable cluster center in common embedding space for its vague and diverse visual region embeddings in different videos. Therefore, we can take the textual embedding of the object as the substitute of every possible object $z$ from different videos in backdoor adjustment:
\begin{equation}
	\begin{aligned}
		P_c(Y|do(V,Q))&=\sum_{z\in Z}^{}P_c(Y|Q,V,z)P(z)\\
		&=\sum_{o_k}^{}P_c(Y|Q,G(V,o_k))P(o_k)\\
		&\overset{\text{\tiny NWGM}}{\approx} P_c(Y|Q,\sum_{o_k}G(V,o_k)P(o_k))
	\end{aligned} \label{eq:backdoor_implement}
\end{equation}
where $G(V,o_k)$ is a fusion module that makes the visual embedding of video $V$ and textual embedding of $o_k$ fully interact with each other in common embedding space via element-wise addition, and $P(o_k)$ is the prior probability of each object $o_k$ appearing in the dataset. Besides, to reduce the expensive cost of separately calculating $P_c(Y|Q,G(V,o_k))$ for every $o_k$, we make an approximation by adopting the Normalized Weighted Geometric Mean (NWGM)~\cite{xu2015show} to absorb the outer sum into the condition variables. Note that the learned causal association $P_c(Y|do(V,Q))$ is also adopted at inference phase, therefore it should be designed as efficient to compute.

\subsection{Model Learning}

We learn the object-relevant association and remove the bad confounding effect from it under the frame-level MIL paradigm for WSVOG, whose corresponding training objectives can be seamlessly integrated with the frame-level MIL loss in~\eqnref{newframeMIL}. Specifically, the spatial-temporal adversarial contrastive loss derived from~\eqnref{contrast} which learns the object-relevant association is formulated as follows:
\begin{equation}
	\begin{aligned}
		\mathcal{L}_{\text{ACL}}=\ell(IE_{[\{\bm{\hat{U}_t^s}\}_{t=1}^T]}-IE_{[\{\bm{\hat{U}_t^c}\}_{t=1}^T]})+\ell(IE_{[\bm{\hat{H}^s}]}-IE_{[\bm{\hat{H}^c}]})\\
	\end{aligned} \label{eq:spatial_temporal_cl}
\end{equation}
where the subscript, for example $[\bm{\hat{H}^c}]$, denotes the intervened value of the proxy variable in $IE(p|do(\bm{H^c}=\bm{\hat{H}^c}))$ in~\eqnref{contrast}. After fully trained using $\mathcal{L}=\mathcal{L}_{\text{MIL}}+\mathcal{L}_{\text{ACL}}$, we learn the object-relevant association $P_c(Y|V,Q)$ in place of $P(Y|V,Q)$. Then we remove the bad confounding effect from it using~\eqnref{backdoor_implement} for the variables and we denote the corresponding loss as $\mathcal{L}'$. By learning the deconfounded object-relevant association $P_c(Y|do(V,Q))$, we are able to achieve an essentially causal WSVOG model, which is endowed with more accurate and robust grounding capability.

\section{Experiment}
\subsection{Dataset}
\noindent \textbf{YouCook2-BB}~\cite{zhou2018weakly}. It is a large-scale dataset consisting of 2,000 YouTube cooking videos from 89 recipes. Each video includes 3 to 15 cooking steps and each step is accompanied with a natural language sentence. A total of 15K pairs of video segments and corresponding sentences are extracted from these steps. The average duration of each video segment is 19.6 seconds and the average length of each sentence is 8.8 words. The training, validation, and testing sets have 9,184, 3,042, and 1,423 pairs, respectively. There are 67 most frequently-appearing objects in the natural language sentences and their bounding boxes are annotated in the validation and testing sets.

\vspace{1em}
\noindent \textbf{ActivityNetEntities}~\cite{zhou2019grounded}. It is a much larger dataset with 52K video segments collected from the ActivityNet-Caption dataset~\cite{krishna2017dense}. Each video segment is paired with at least one noun phrase in the natural language sentence. The training and validation sets have 34,059 and 8,468 pairs, respectively. The annotations of testing set are not public, and we follow~\cite{yang2020weakly} to evaluate our model on the validation set. There are 432 most frequently-appearing objects and their bounding boxes are annotated in the training and validation sets. We do not use the annotations in the training set under our weakly-supervised setting.

\begin{table*}[t]
	
	\label{exp-table1}
	\setlength{\tabcolsep}{15pt}
	\caption{Weakly-supervised video object grounding results on YouCook2-BB. }\label{tab:YC2}
	\vspace{-4mm}
	\begin{center}
		{	
			\renewcommand{\arraystretch}{1.4} % default is 1.0
			\begin{tabular}{l | c c| c c  || c c| c c  }
				\Xhline{1pt}
				\multicolumn{1}{l|}{\multirow{3}{*}{Methods}} &  \multicolumn{4}{c||}{{Query Accuracy (\%)}} & \multicolumn{4}{c}{{Box Accuracy (\%)}}  \\
				
				\cline{2-9}
				& \multicolumn{2}{c| }{macro} & \multicolumn{2}{c|| }{micro} & \multicolumn{2}{c| }{macro} & \multicolumn{2}{c }{micro} \\
				\cline{2-3}
				\cline{4-5}
				\cline{5-6}
				\cline{7-9}
				& val & test & val & test & val & test & val & test\\
				\Xhline{1pt}
				Upper Bound   &65.55 &65.72 &70.32 &70.24 &62.42 &64.41 &68.56  &68.74\\
				\hline
				Random  &11.15 &10.94 &12.96 &12.79 & 10.36 & 10.67 &12.19 &12.26  \\
				Extended DVSA~\cite{karpathy2015deep}  &38.20 &37.98  &45.60 &44.79 & 36.67 & 36.30 &43.62 &42.87\\
				VOG~\cite{zhou2018weakly}  &37.26 &36.69  &44.99 &44.34 & 35.69 & 35.08 &43.04 &42.42 \\
				NAFAE~\cite{shi2019not}  &41.29 &42.45  &48.52 &48.41 & 39.54 & 40.71 &46.41 &46.33  \\
				Chen \emph{et al.}~\cite{chen2020activity}   &41.43 &42.55 &49.71 &48.91  & 40.66 & 41.67 &\textbf{49.11} &48.22 \\
				STVG~\cite{yang2020weakly} (baseline)   &41.36 &43.40 &48.74 &48.98 & 39.90 & 41.63 &46.80 &47.02  \\
				\hline
				\textbf{Ours}  &\textbf{43.42} &\textbf{44.51} &\textbf{50.51} &\textbf{50.62}  &\textbf{41.93} &\textbf{42.71} & {48.51} & \textbf{48.61} \\
				\Xhline{1pt}
			\end{tabular} 
		}
	\end{center}
	\vspace{-3mm}
	
\end{table*}

\begin{table}[htbp]
	\label{exp-table1}
	\setlength{\tabcolsep}{9pt}
	\centering
	\caption{WSVOG results on ActivityNetEntities.} \label{tab:Anet}
	\vspace{-4mm}
	\begin{center}
		{	
			\renewcommand{\arraystretch}{1.4} % default is 1.0
			\begin{tabular}{l | c| c|| c| c}
				\Xhline{1pt}
				\multicolumn{1}{l|}{\multirow{2}{*}{Methods}} &  \multicolumn{2}{c||}{{Query Acc. (\%)}} & \multicolumn{2}{c}{{Box Acc. (\%)}}  \\
				
				\cline{2-5}
				& macro & micro & macro & micro \\
				\Xhline{1pt}
				Upper Bound & 50.94 & 64.81 &49.05 &63.41 \\
				\hline
				Random &3.29 &4.81 &3.16 &4.03 \\
				NAFAE~\cite{shi2019not} &21.47 &32.06 &19.54 &28.08 \\
				STVG~\cite{yang2020weakly} &21.93 &34.00  &20.00 &29.78\\
				$\text{STVG}^*$ (baseline)  &23.01 &29.09  &21.12 &25.81\\
				\hline
				\textbf{Ours}  &\textbf{29.36} &\textbf{37.89} &\textbf{27.13} &\textbf{33.55} \\

				\Xhline{1pt}
			\end{tabular} 
		}
	\end{center}
	\vspace{-4mm} 
	
\end{table}

\vspace{1em}
\noindent \textbf{RoboWatch}~\cite{sener2015unsupervised}. It has 225 YouTube videos having no overlap with YouCook2-BB. Each video segment is accompanied with a natural language sentence. \cite{huang2018finding} extended the bounding box annotations for part of these videos. Most of the objects appeared in RoboWatch have not occurred in YouCook2-BB. We conduct Out-Of-Distribution (OOD) test on it to evaluate whether our model truly learns the causal association, by first pretraining the model on the training set of YouCook2-BB and directly evaluating it on RoboWatch.

\vspace{-1em}
\subsection{Evaluation Metric}
We follow the same evaluation metric employed in previous works~\cite{shi2019not,yang2020weakly,chen2020activity}. The overall performance is evaluated using Box Accuracy and Query Accuracy. Box Accuracy is defined as the ratio of correctly grounded boxes among all the grounded boxes. Specifically, the top-1 ranked box in each frame is selected as the grounded box, and we treat the box having more than 50\% Intersection-of-Union (IoU) with the groundtruth as correct. Query Accuracy is defined as the ratio of correctly grounded queries to all of the queries. A grounded query is correct if it matches with the correctly grounded box, where query is the sentence describing objects. To further evaluate the averaging accuracy for all the classes and the global accuracy regardless of distinct classes, we additionally \mbox{adopt \emph{macro-accuracy} and \emph{micro-accuracy}, respectively.}

\vspace{-1em}
\subsection{Implementation Details} \label{sec:implementation}
We adopt STVG~\cite{yang2020weakly} as the backbone of the frame-level MIL framework described in~\secref{MIL}, and we follow the same settings of STVG for fair comparison. Note that the paradigm of frame-level MIL framework is standardly adopted in existing WSVOG methods, which is not our novelty in this paper. For each video segment, we uniformly sample $T=5$ frames in YouCook2-BB dataset and $T=6$ in ActivityNetEntities dataset for training. The region proposals in YouCook2-BB and RoboWatch are extracted following the same procedure of Shi \emph{et al.}~\cite{shi2019not} that utilizes Faster R-CNN~\cite{ren2016faster} pretrained on Visual Genome~\cite{krishna2017visual} with VGG-16 backbone, and the extracted region feature dimension is 4096. For ActivityNetEntities, we directly employ the extracted 2048-dimensional region proposals provided by~\cite{zhou2019grounded}, which have been also adopted in other compared methods. We select top $N=20$ region proposals in each frame. For objects described in the sentence, we employ GloVe~\cite{pennington2014glove} as the textual features, whose dimension is 300. In STVG backbone, we mainly learn two embedding modules, namely visual embedding module $E_v(\cdot)$ and textual embedding module $E_q(\cdot)$, to embed visual region features and textual object features into the 512-dimensional common embedding space, where similarities can be directly measured. 

For the implementation of learning object-relevant association, we select top $B_r=10$ regions in each frame-bag for region-level \emph{content} set and top $B_f=1$ frame in the video for frame-level \emph{content} set. For adversarial sample generation, we construct region- and frame-level memory banks containing $J=100$ and $W=1000$ memory vectors, respectively. In adversarial contrastive learning, the intervened sizes of \emph{content} set and \emph{style} set are set to the same by adopting the smaller size of them. In terms of memory bank update, we set the momentum factor $\alpha$ to $0.9$. The adversarial contrastive learning is performed separately in spatial and temporal because they will affect each other. For removing the confounding effect using backdoor adjustment, we implement the fusion module $G(V,o_k)$, which is supposed to be efficient to compute, as $G(V,o_k)=E_v(V)\oplus E_q(o_k)$ where $\oplus$ is the element-wise broadcast add operator and $E_v(V)$ is equivalent to the set of $\{E_v(r_n^t)\}$. We have also tried other attention-based fusion strategies~\cite{wang2020visual,yang2021deconfounded}, but found this design simple and effective, which saves much computational overhead. The scale parameter $\eta$ in $\ell(x)$ is set to 0.2 for YouCook2-BB and 0.05 for ActivityNetEntities. For these two datasets, the batchsizes are set to 48, 16, and the learning rates are set to 1e-4, 8e-4, respectively. We implement it using PyTorch and we \mbox{adopt Adam optimizer with 1e-4 weight decay. }

\subsection{Comparison with State-of-the-Arts}
We compare our method with existing state-of-the-arts on YouCook2-BB and ActivityNetEntities datasets. We outperform all of them by a considerable margin, which proves the effectiveness of our method.

\vspace{0.3em}
\noindent \textbf{Compared Methods.} The following five existing methods related to WSVOG task are compared. Note that this task is still being explored in the research community, and we have listed all the existing state-of-the-arts for comparison.

\begin{myitemize2}
	\item \textbf{Extended DVSA~\cite{karpathy2015deep}:} It is an extension of the visual grounding method in image domain. It is the only one in compared methods that does not adopt the frame-level MIL framework but takes an entire video as a video-bag.
	\item \textbf{VOG~\cite{zhou2018weakly}:} It is the first method that specially tackles the WSVOG task, which adopts the frame-level MIL framework with loss weighting and object interaction.
	\item \textbf{NAFAE~\cite{shi2019not}:} It adopts a better weighting strategy for each frame-bag, and enhances visual representations by clustering similar visual objects across frames.
	\item \textbf{Chen \emph{et al.}~\cite{chen2020activity}:} It enhances visual and textual representations by exploiting activity cues with the help of external human pose detector and large-scale pretrained language model.
	\item \textbf{STVG~\cite{yang2020weakly}:} It simultaneously models the contextual similarities within regions in each spatial frame-bag and frames across the temporal aggregation of the video in an end-to-end manner.
\end{myitemize2}

For fair comparison, all the compared methods utilize the same visual region proposals, and we directly cite their numbers from original papers. Note that it is not a completely fair comparison with Chen \emph{et al.}'s work~\cite{chen2020activity}, because they utilize extra information from external human pose detectors and large-scale pretrained language models to enhance the feature representations. Aside from the above five state-of-the-art methods, we additionally provide Random results as a baseline, and Upper Bound indicates how far these methods are from the performance limit~\cite{zhou2018weakly}.

\vspace{0.3em}
\noindent \textbf{YouCook2-BB.} The results on YouCook2-BB dataset are summarized in~\tabref{YC2}. Our method outperforms all the compared state-of-the-arts by a considerable margin in almost all evaluation metrics, which proves the effectiveness of our method. In particular, we surpass the strong baseline STVG~\cite{yang2020weakly} up to 2.06\% absolute gains and averagely 1.62\% absolute gains. We consistently surpass Chen \emph{et al.}'s work~\cite{chen2020activity} significantly despite that they resort to external information for better feature representations, which shows the great potential of our method. It is also worth noting that recent works focusing on either designing more delicate architectures or enhancing feature representations using external information have only achieved relatively minor improvements, due to the bottleneck of this dataset. In contrast, our method boosts the performance significantly merely by using the power of causal analysis which aims to learn the deconfounded object-relevant association instead of previous statistical association from the training data.

\vspace{0.3em}
\noindent \textbf{ActivityNetEntities.} The compared state-of-the-arts are NAFAE~\cite{shi2019not} and STVG~\cite{yang2020weakly} that have reported their performance on this dataset. We retrain the STVG by ourselves since they did not provide the pretrained model or code on it, and we denote our re-implemented baseline model as $\text{STVG}^{*}$. Results are summarized in~\tabref{Anet}, from which we can observe that our model consistently outperforms all the compared state-of-the-arts by a large margin in all evaluation metrics. Specifically, we surpass our re-implemented $\text{STVG}^{*}$ up to 8.34\% absolute gains, and we also surpass the originally reported results of STVG up to 7.92\% absolute gains. Besides, we can see that our model obtains more significant gains on this dataset than on YouCook2-BB, because ActivityNetEntities is a much larger dataset consisting of more video segments and more types of objects. Thus, the compared methods will suffer from more severe ambiguity under weak supervision as well as the severe confounding effect brought by the diversely co-occurred objects in ActivityNetEntities. Our model can effectively alleviate these problems by pursuing the true causality within the association, thus obtaining much better grounding performance.

\vspace{-0.5em}
\begin{table}[htbp]
	\label{exp-table1}
	\setlength{\tabcolsep}{8pt}
	\centering
	\caption{Out-Of-Distribution (OOD) test on RoboWatch.} \label{tab:robowatch}
	\vspace{-4mm}
	\begin{center}
		{	
			\renewcommand{\arraystretch}{1.4} % default is 1.0
			\begin{tabular}{l | c}
				\Xhline{1pt}
				\multicolumn{1}{l|}{\multirow{1}{*}{Methods}} &  \multicolumn{1}{c}{{query micro-acc (\%)}}  \\
				\Xhline{1pt}
				Random &8.03 \\
				Extended DVSA~\cite{karpathy2015deep}  &28.25 \\	
				NAFAE~\cite{shi2019not}  &31.68 \\
				$\text{STVG}^*$~\cite{yang2020weakly} (baseline)  &32.05 \\
				\hline
				\textbf{Ours}  &\textbf{36.12} \\
				
				\Xhline{1pt}
			\end{tabular} 
		}
	\end{center}
	\vspace{-6mm}
	
\end{table}

\subsection{Out-Of-Distribution (OOD) Test}
In the above evaluations, the split of training, validation, and testing sets follows the Independent and Identically Distributed (IID) assumption~\cite{vapnik1999overview}. While the IID setting is commonly adopted in WSVOG task, it is not sufficient to evaluate whether the causality is actually learned in our model, due to the consistently existed biases during learning and inference. To this end, we introduce the Out-Of-Distribution (OOD) test~\cite{teney2020value} into evaluation. Specifically, we first train the model on the training set of YouCook2-BB, then we evaluate it on the testing set of RoboWatch. RoboWatch is differently distributed in that it includes different types of videos having no overlap with YouCook2-BB and it consists of many objects that have never occurred in YouCook2-BB, such as ``tie" and ``hanger".

\begin{table*}[t]
	
	\label{exp-table1}
	\setlength{\tabcolsep}{8pt}
	\caption{Ablation results on YouCook2-BB and corresponding OOD test on RoboWatch.}\label{tab:ablation_YC2}
	\vspace{-4mm}
	\begin{center}
		{	
			\renewcommand{\arraystretch}{1.4} % default is 1.0
			\begin{tabular}{l | c c| c c  || c c| c c ||| c  }
				\Xhline{1pt}
				\multicolumn{1}{l|}{\multirow{4}{*}{Methods}} &  \multicolumn{8}{c|||}{{\textbf{YouCook2-BB}}} &  \multicolumn{1}{c}{{\textbf{RoboWatch}}} \\
				\cline{2-10}
				&  \multicolumn{4}{c||}{{Query Accuracy (\%)}} & \multicolumn{4}{c|||}{{Box Accuracy (\%)}} &  \multicolumn{1}{c}{{Query Acc. (\%)}}  \\
				
				\cline{2-10}
				& \multicolumn{2}{c| }{macro} & \multicolumn{2}{c|| }{micro} & \multicolumn{2}{c| }{macro} & \multicolumn{2}{c |||}{micro}  & micro  \\
				\cline{2-10}
				& val & test & val & test & val & test & val & test & OOD~~test \\
				\Xhline{1pt}
				STVG~\cite{yang2020weakly} (baseline)   &41.36 &43.40 &48.74 &48.98 & 39.90 & 41.63 &46.80 &47.02  & 32.05 \\			
				\hline
				\textbf{Learning Object-Relevant Association}   &  &  &  &  &   &   &  &  &   \\
				Baseline + ACL in Spatial    &41.93 &43.55 &49.83 &49.54 & 40.47 & 41.81 &47.84 &47.57  & 35.09  \\
				Baseline + ACL in Temporal   &42.75 &43.69 &49.90 &49.86 & 41.28 & 41.95 &47.92 &47.88  & 34.45  \\
				Baseline + ACL   &42.81 &44.36 &50.35 &50.45 & 41.32 & 42.57 &48.34 &48.45  & 35.47  \\
				Full w/o ACL  &42.66  &43.71  &49.77  &49.52  &41.19   &41.98   &47.80  &47.56  & 35.89\\
				\hline
				\textbf{Removing Bad Confounding Effect}   &  &  &  &  &   &   &  &   &   \\
				Full w/o BDA   &42.81 &44.36 &50.35 &50.45 & 41.32 & 42.57 &48.34 &48.45  & 35.47  \\
				Baseline + BDA  &42.66  &43.71  &49.77  &49.52  &41.19   &41.98   &47.80  &47.56  & 35.89\\
				Full w/ attention-based $G(\cdot)$   &42.65  &44.14  &50.32  &50.29  &41.16   &42.37   &48.32  &48.29  & 35.77  \\
				\hline
				\textbf{Full Model} &  &  &  &  &   &   &  &   &   \\
				Baseline + ACL + BDA  &\textbf{43.46} &\textbf{44.61} &\textbf{50.45} &\textbf{50.61}  &\textbf{41.94} &\textbf{42.80} & \textbf{48.46} & \textbf{48.60} &\textbf{36.12} \\
				\Xhline{1pt}
			\end{tabular} 
		}
	\end{center}
	\vspace{-5mm}
	
\end{table*}

\begin{table}[htbp]
	\label{exp-table1}
	\setlength{\tabcolsep}{6pt}
	\centering
	\caption{Ablation results on ActivityNetEntities.} \label{tab:ablation_ANET}
	\vspace{-4mm}
	\begin{center}
		{	
			\renewcommand{\arraystretch}{1.4} % default is 1.0
			\begin{tabular}{l | c| c|| c| c}
				\Xhline{1pt}
				\multicolumn{1}{l|}{\multirow{2}{*}{Methods}} &  \multicolumn{2}{c||}{{Query Acc. (\%)}} & \multicolumn{2}{c}{{Box Acc. (\%)}}  \\
				
				\cline{2-5}
				& macro & micro & macro & micro \\
				\Xhline{1pt}
				$\text{STVG}^*$~\cite{yang2020weakly} (baseline)  &23.01 &29.09  &21.12 &25.81\\
				\hline
				\textbf{Ablated Versions}  & & & & \\
				Baseline + ACL in Spatial &28.10 &36.51  &25.89 &32.16 \\
				Baseline + ACL in Temporal &26.17 &31.24  &23.95 &27.69\\
				Baseline + ACL   &28.58 &36.75  &26.32 &32.44\\
				Baseline + BDA &27.30 &31.45 &24.99 &27.71 \\
				\hline
				\textbf{Full Model}  & & & & \\
				Baseline + ACL + BDA  &\textbf{29.36} &\textbf{37.89} &\textbf{27.13} &\textbf{33.55}  \\

				\Xhline{1pt}
			\end{tabular} 
		}
	\end{center}
	\vspace{-6mm}
	
\end{table}

\vspace{0.3em}
\noindent \textbf{RoboWatch.} For fair comparison, all the compared methods are trained and evaluated using the same visual and textual features. $\text{STVG}^*$ denotes our reported results of baseline STVG on OOD test. Results are shown in~\tabref{robowatch}, from which we can see that our method outperforms other state-of-the-arts by quite a large margin, with more than 4\% absolute gains. It is important to point out that the relative improvements of our model on OOD test are significantly larger than those on IID test (which are shown in~\tabref{YC2}). Because the compared methods which ignore the pursuit of true causality but learn the statistical association during training are vulnerable to the biases in datasets and sensitive to the distribution changes, thus leading to only limited improvements on OOD test. In contrast, we resort to the causal analysis to force the model to learn object-relevant association and remove the bad confounding effect from it, which is robust to distribution changes on OOD test.

\subsection{Further Remarks}

\vspace{-0.3em}
\noindent \textbf{Effectiveness of learning object-relevant association.} We provide four types of ablated versions to evaluate its effectiveness. Results are shown in~\tabref{ablation_YC2}, where we denote Learning Object-Relevant Association as ACL. Specifically, we first conduct ACL on our baseline STVG in spatial and temporal, respectively. It can be seen that our adversarial contrastive learning framework can mitigate the ambiguity, which is intrinsically possessed under weak supervision, in both spatial and temporal levels by forcing the learned association to satisfy its object-relevant interventional effect. Moreover, they can cooperate spatial-temporally to achieve a better performance. We also remove ACL from the full model to evaluate its contribution to the overall performance, and we observe a significant decrease in performance, proving the effectiveness of learning object-relevant association. We additionally take the similar evaluations on another dataset ActivityNetEntities, and results shown in~\tabref{ablation_ANET} consistently prove its effectiveness. We also notice that ACL in Temporal brings less improvements than ACL in Spatial on ActivityNetEntities. The reason is that this dataset suffers from less temporal ambiguity led by frequent shot cut in videos, since these videos are mainly about temporally consistent activities \mbox{rather than cooking videos with frequent shot cut.}

\vspace{0.3em}
\noindent \textbf{Effectiveness of removing bad confounding effect.} To evaluate the effectiveness of removing bad confounding effect from the learned association, we wipe out this module (denoted as BDA) from the full model and the corresponding performance consistently drops as shown in~\tabref{ablation_YC2}. We also investigate its potential by directly applying the backdoor adjustment on the baseline STVG and we find that it can significantly improve the performance as well, which we will explain in the next paragraph. For the implementation of fusion module $G(\cdot)$ in backdoor adjustment, we have also tried another popular attention-based implementation~\cite{wang2020visual} but found it is not so effective in WSVOG task. The reason may lie in that visual regions are extremely spatial-temporally noisy and diverse in WSVOG and the attention-based fusion strategy cannot effectively capture the meaningful relations. In contrast, our element-wise addition for $G(\cdot)$ is simple and stable for training, thus showing a superior performance. Besides, please note that in~\tabref{ablation_YC2}, Full w/o ACL is equivalent to Baseline + BDA, and Full w/o BDA is equivalent to \mbox{Baseline + ACL just for convenient reference.}

\vspace{0.3em}
\noindent \textbf{Are the above two modules redundant?} Astute readers who are knowledgeable in causality may think that in our unified causal graph in~\figref{causal_graph}, if we intervene the value of $V$, the input link $Z\rightarrow V$ will be cut off since $V$ will only be controlled by our intervention. Therefore, there is no need to conduct backdoor adjustment to remove the confounding effect brought by $Z$. They may also think that if we directly remove the confounder (maybe not objects but contexts) via backdoor adjustment at the very beginning, the model will learn the purely causal association $V\rightarrow Y$, so that learning object-relevant association afterwards is unnecessary. For the first concern, we have to point out that in our adversarial contrastive learning, we only intervene part of $V$ from the perspective of data generation and we do not completely wipe out its value but assign it with the adversarial feature. Therefore, the link $Z\rightarrow V$ still holds, leading to the confounding effect. Moreover, from the ablated results of Baseline + BDA and Full w/o BDA in~\tabref{ablation_YC2}, we can observe that the backdoor adjustment brings more gains on baseline than on the object-relevant association, indicating that the link $Z\rightarrow V$ becomes weaker after adversarial contrastive learning but it still exists. For the second concern, we have to admit that it does hold for fully-supervised learning, since there is no ambiguity problem under explicit strong supervision signals. However, under weak supervision in WSVOG, the learned association is not just simply confounded by a certain confounder, but the association itself is extremely ambiguous. Under such circumstances, the intrinsically possessed spatial-temporal ambiguity in $V$ cannot be well mitigated via only backdoor adjustment which aims to remove the confounding effect. Hence, we resort to interventional effect to force the model to focus on the causal factors among the ambiguous association. Aside from the above theoretical analysis, the ablated results also prove that the two modules are not redundant.

 \vspace{0.3em}
 \noindent \textbf{Why not resort to other causal effect like Total Direct Effect (TDE)?} We observe that there are some recent works pursuing the causal effect of Total Direct Effect (TDE)~\cite{tang2020unbiased}, Total Indirect Effect (TIE)~\cite{wang2021clicks}, and so on. The key idea behind them is to remove the unwanted effect that will lead to shortcut bias in learning and inference, by subtracting the unwanted predictions and retaining the wanted ones. Typically, they obtain unwanted predictions by intervening the certain input to take the value of zeros or constants, so as to see what the model will predict if nothing is input. Besides, by pursuing causal effect like TDE, they will not be affected by confounders any more, because they make a thorough intervention by directly manipulating the value to be zeros, cutting off the link $Z\rightarrow V$ completely. In our WSVOG task, it seems that we can also pursue the Total Direct Effect (TDE) by intervening the \emph{content} $C$ to be zeros so as to see what the model will predict if only the \emph{style} $S$ is seen, then subtracting this obtained predictions from the total predictions. However, the key problem here is that we cannot effectively distinguish \emph{content} $C$ and \emph{style} $S$ since there is no strong supervision for us to tell them apart in weakly-supervised setting. Therefore, we propose to use adversarial contrastive learning for the obtained causal effect instead of simply subtracting the zero-input predictions in TDE.

 \vspace{0.3em}
\noindent \textbf{OOD test for ablated models.} To better evaluate whether the true causalities are learned in our model, we use the ablated models pretrained on YouCook2-BB to take the OOD test on RoboWatch. Results are illustrated in~\tabref{ablation_YC2}. We can see that either learning the object-relevant association or removing the bad confounding effect can boost the OOD test performance significantly. 
Besides, it is observed that the OOD improvement of BDA is typically larger than that of ACL, indicating that removing the bad confounding effect via backdoor adjustment can better adapt to distribution changes for different datasets.

 \vspace{0.3em}
 \noindent \textbf{Where and when to perform the backdoor adjustment?} In our baseline STVG frame-level MIL framework, the visual embedding module $E_v(\cdot)$ is made up of a self-attention component for visual regions in each spatial frame-bag. It seems that we can perform the backdoor adjustment using $G(\cdot)$ for visual features either before or after the self-attention module. However, in causal analysis, the self-attentive operation on features can be viewed as an indiscriminate correlation against the \emph{do}-calculus~\cite{wang2020visual}. Hence, we adopt $G(\cdot)$ for visual features after self-attention module. In terms of when to perform the backdoor adjustment, readers may concern that after the model is fully trained using $\mathcal{L}=\mathcal{L}_{\text{MIL}}+\mathcal{L}_{\text{ACL}}$, conducting $G(\cdot)$ for the learned visual features will change their feature distribution and thus harm the model. In fact, we train the model with $\mathcal{L}$ and $\mathcal{L}'$ separately and make residual connections for these learned features at test phase. Similar operations can also be found in~\cite{wang2020visual}, where the extracted features after backdoor adjustment are viewed as preserving the visual causality and should be cooperated with the original features for maximum utilization.
 
   \begin{figure}[tbp]
	\centering
	\includegraphics[width=0.78\columnwidth]{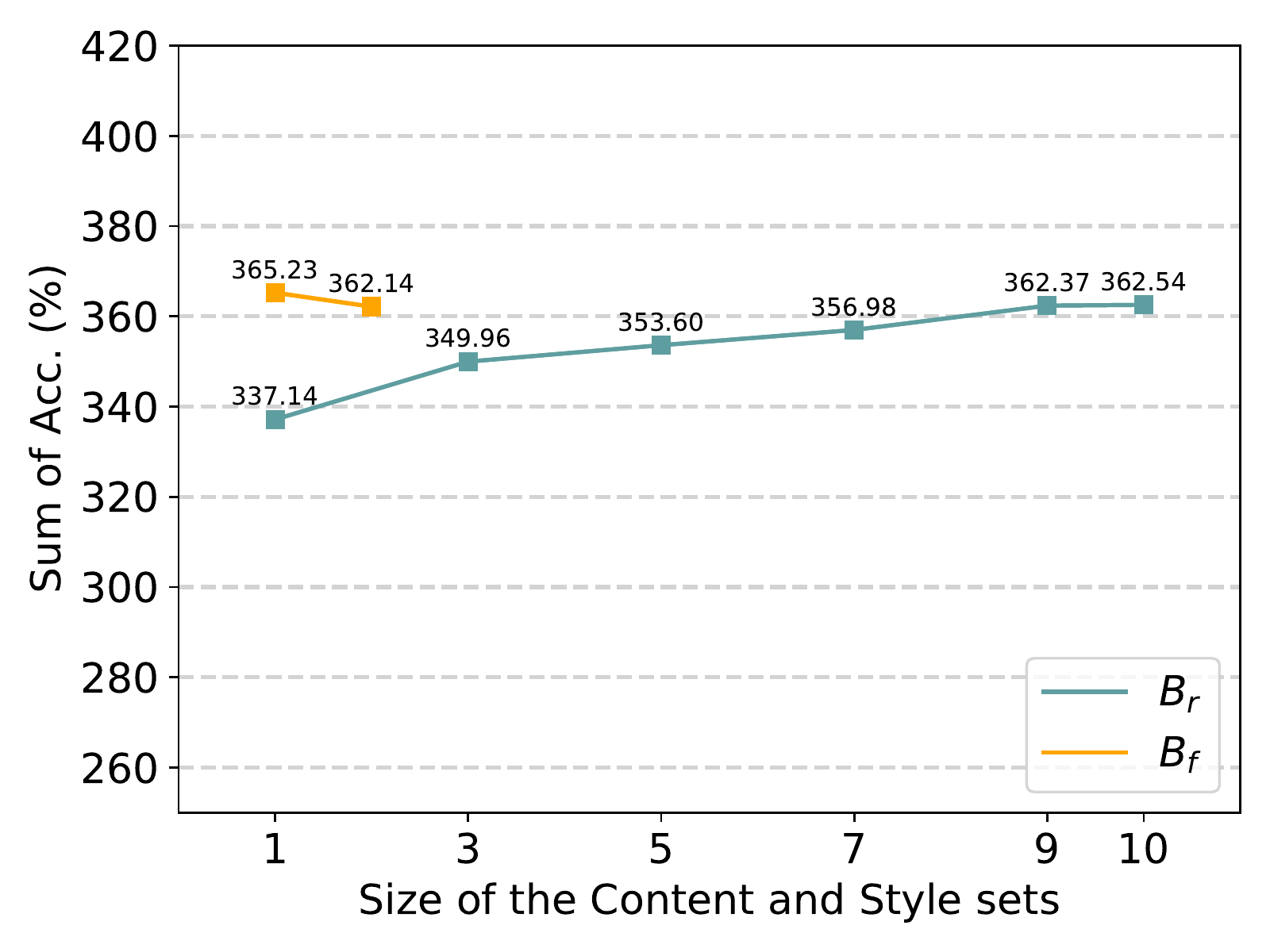}
	\vspace{-4mm}
	\caption{Effect of the sizes of \emph{content} set and \emph{style} set, which are $B_r$ and $B_f$, respectively.} \label{fig:size_set}
	\vspace{-5mm}
\end{figure}

\vspace{-1.3em}
\begin{figure}[hbp]
	\centering
	\includegraphics[width=0.78\columnwidth]{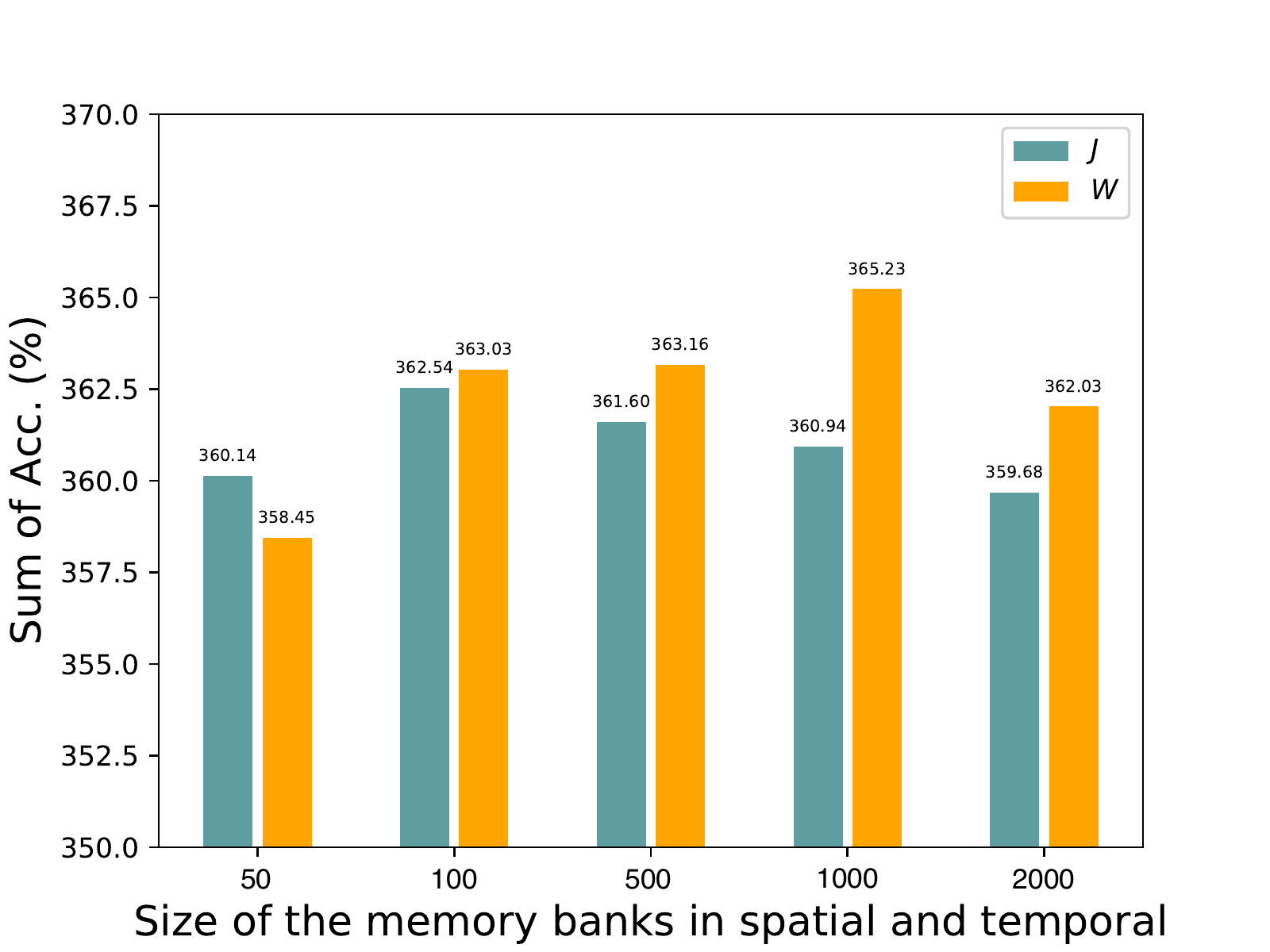}
	\vspace{-3mm}
	\caption{Effect of the sizes of the memory banks in spatial and temporal, which are $J$ and $W$, respectively.} \label{fig:size_mem}
	\vspace{-2mm}
\end{figure}

 \begin{figure*}[thbp]
	\centering
	\includegraphics[width=\linewidth]{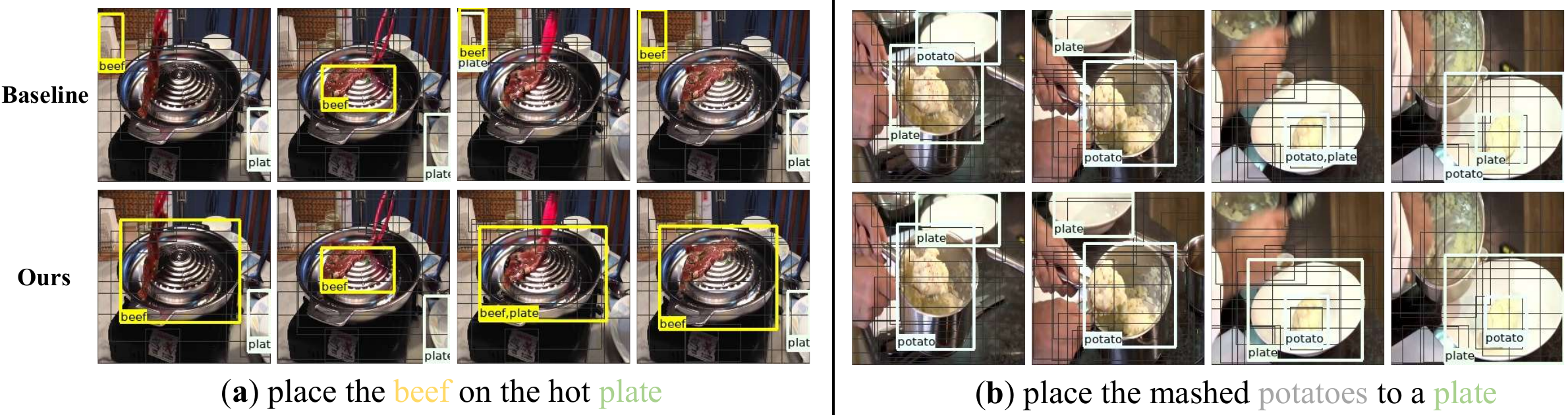}
	\vspace{-6mm}
	\caption{Two qualitative grounding results where: (a) shows the ambiguity problem under weak supervision and we mitigate it by learning object-relevant association; (b) shows the confounding effect brought by co-occurred objects and we remove such effect using backdoor adjustment.} \label{fig:QA_all}
	\vspace{-2mm}
\end{figure*} 

\begin{figure*}[thbp]
	\centering
	\includegraphics[width=\linewidth]{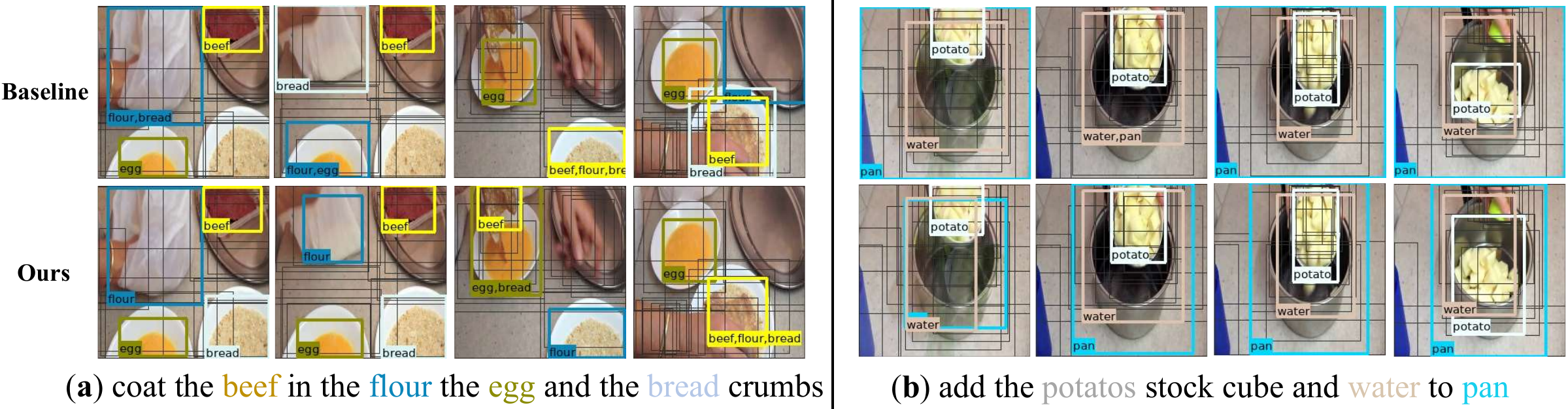}
	\vspace{-6mm}
	\caption{More qualitative grounding results for reference, from which we can see that our model consistently shows the better grounding performance.} \label{fig:QA_other_all}
	\vspace{-3mm}
\end{figure*} 

\begin{figure}[thbp]
	\centering
	\includegraphics[width=\columnwidth]{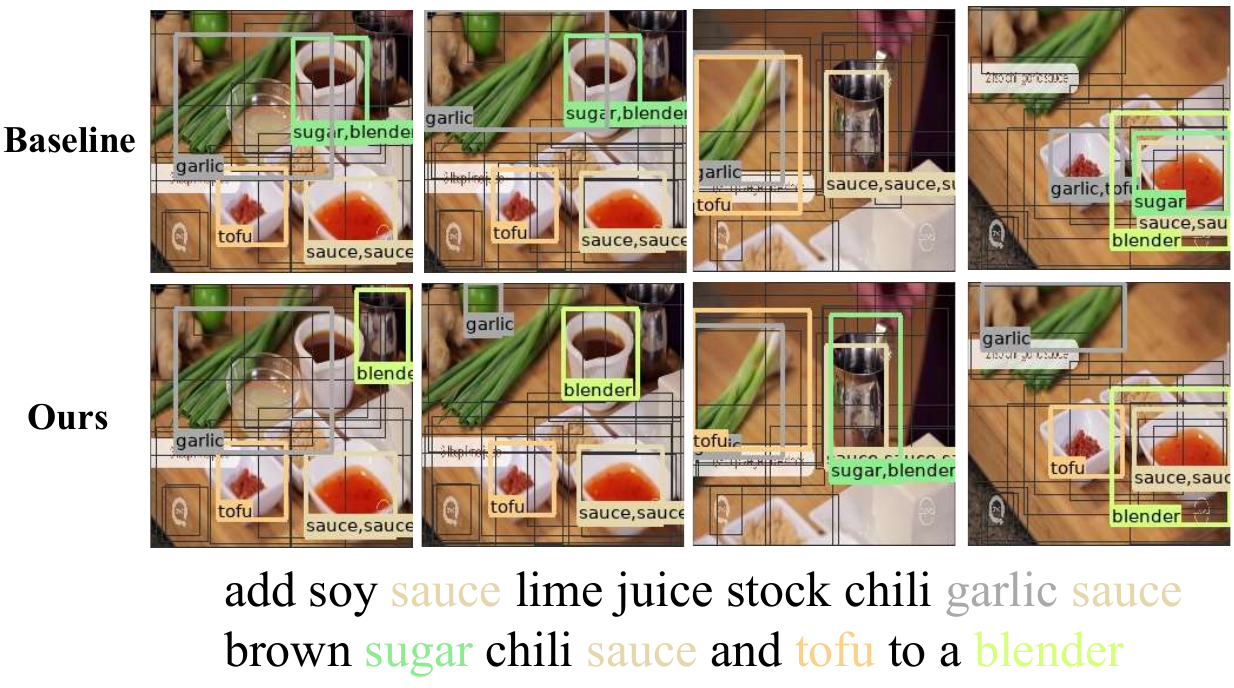}
	\vspace{-7mm}
	\caption{Failure cases of our model, which are mainly resulted from the object occlusion, object out of scene, and small object.} \label{fig:QA_fail}
	\vspace{-6mm}
\end{figure} 
 
 \vspace{0.3em}
 \noindent \textbf{Effect of the sizes of \emph{content} set and \emph{style} set.} As we have described in~\secref{implementation} that in adversarial contrastive learning, the intervened sizes of \emph{content} $C$ and \emph{style} $S$ are set to the same by adopting the smaller size of them. When performing adversarial contrastive learning in spatial (or temporal), the shared size of the two sets is $B_r$ (or $B_f$). On YouCook2-BB dataset, their values are set to $B_r=10$ and $B_f=1$. We conduct sensitivity analysis on Baseline + ACL in Spatial (or Temporal) to evaluate their effect. Note that $B_r$ and $B_f$ are both no larger than half of the visual regions in each frame ($B_r\leq \lfloor20/2\rfloor=10$) and half of the sampled frames in each video ($B_f\leq\lfloor5/2\rfloor=2$), or we cannot balance the sizes of \emph{content} $C$ and \emph{style} $S$. Results are illustrated in~\figref{size_set}. We can see that in spatial, the performance is positively related to the size $B_r$. We infer the reason is that although there are only limited positive visual regions (proxy variables of \emph{content} $C$) in each spatial frame-bag, we are not able to accurately select them out based on spatial grounding score $M^k_{t,n}$ in practice, due to the weak supervision. Therefore, we have to enlarge the size of \emph{content} set so as to include more potential proxy variables of \emph{content} $C$ for better adversarial contrastive learning. In contrast, we find that in temporal, the performance of smaller size of sets ($B_f=1$) is better than that of $B_f=2$. It is interesting to explain such difference, which is mainly due to the well-known \emph{false-positive} problem commonly pointed out by other existing works~\cite{shi2019not,yang2020weakly}. To be specific, a frame-bag can simply have no positive visual region in it because objects are not guaranteed to appear in every frame of the video. But a video must have at least one frame containing the described object (which is guaranteed by the video-level annotation). Therefore, in temporal, the top-ranked frames are more likely to be the proxy variables of \emph{content} $C$, thus a smaller size of sets can work well.

\noindent \textbf{Effect of the sizes of memory banks.} In adversarial contrastive learning, we generate adversarial samples based on the memory banks. We would like to see the effect of the sizes of memory banks, hence we conduct the sensitivity analysis on Baseline + ACL in Spatial (or Temporal). As illustrated in~\figref{size_mem}, the sizes of memory banks in spatial and temporal are denoted as $J$ and $W$, respectively. It is observed that the optimal performance in spatial is achieved with the size of $J=100$, and that in temporal is achieved with the size of $W=1000$. Too small or too large sizes of the memory banks will both harm the performance, because small memory banks cannot provide enough information for adversarial sample generation and large ones bring too much noise. Besides, a larger memory bank is needed in temporal than that in spatial, because the frame-level memory bank in temporal has to accommodate numerous combinations of objects that may appear in a frame, while the region-level memory bank in spatial is supposed to accommodate limited types of visual objects or background.

\vspace{-1em}
\subsection{Qualitative Results}
We present some grounding results of our proposed method to qualitatively demonstrate its effectiveness. We respectively show how our model learns object-relevant association and how it removes the bad confounding effect. We also provide additional grounding results and failure cases for reference. All the grounding results are presented \mbox{on the testing set of YouCook2-BB.}

\vspace{0.3em}
\noindent \textbf{Learning object-relevant association.}
As shown in~\figref{QA_all} (a), the baseline model suffers from the ambiguity problem that the grounded visual regions of ``beef" are distracted by other proposals in the background, thus leading to incorrect grounding results (shown in yellow boxes of Baseline). In our model, we learn the object-relevant association by adversarial contrastive learning using Interventional Effect, thus we mitigate such ambiguity in our grounding results where the grounded visual regions focus on the object of ``beef" rather than distracting regions in the background.

\vspace{0.3em}
\noindent \textbf{Removing bad confounding effect.} In~\figref{QA_all} (b), we present a case reflecting the confounding effect, where the visual regions of ``potato" are mistakenly associated with the described object ``plate", due to their frequent co-occurrences. In other words, the baseline model hesitates about the region-object correspondence when encountered with the objects of  ``potato" and ``plate". In our model, we remove such confounding effect and pursue the true causality by adopting backdoor adjustment, and we show a better region-object correspondence in our grounding results.

\vspace{0.3em}
\noindent \textbf{Additional grounding results and failure case.} We additionally provide more grounding results in~\figref{QA_other_all} for reference, from which we can see our model consistently shows a better grounding performance. We also provide the common failure cases in~\figref{QA_fail}, mainly resulted from the object occlusion, object out of scene, and small object. Despite of it, our model still shows its potential to improve the grounding performance (\emph{e.g.,} the grounded boxes of ``blender" and ``garlic" are corrected by our model).

%\vspace{-1em}
\section{Conclusion}
	This paper proposes a unified causal framework for WSVOG, which learns the deconfounded object-relevant association $P_c(Y|{do}(V,Q))$ via causal intervention. By taking a causal view of WSVOG task, we clearly reveal the exact causal paths that lead to the ambiguity problem and confounding effect, and we work out the fundamental solutions to them based on causal intervention. Accordingly, we learn the object-relevant association $P_c(Y|V,Q)$ by causal intervention from the perspective of video data generation process. To overcome the problems of lacking fine-grained supervision in terms of intervention, we propose a novel spatial-temporal adversarial contrastive learning paradigm. To further remove the accompanying confounding effect in $P_c(Y|V,Q)$, which is caused by co-occurred objects, we pursue the true causality $P_c(Y|{do}(V,Q))$ by conducting causal intervention via backdoor adjustment. Finally, the deconfounded object-relevant association $P_c(Y|{do}(V,Q))$ is learned and optimized under the unified causal framework in an end-to-end manner. Extensive experiments on both IID and OOD testing sets demonstrate its accurate and robust grounding performance against state-of-the-arts.

We believe this work could shed light on exploring the new boundary of the causal analysis in computer vision tasks under weak supervision. Unlike previous methods that simply attribute all the spurious association to a certain confounder, we carefully pinpoint the spurious association from two aspects: (1) the association itself is not object-relevant but extremely ambiguous due to weak supervision, and (2) the association is unavoidably confounded by the observational bias when taking the statistics-based matching strategy in existing methods. Accordingly, we propose a unified causal framework, in which the above two problems are clearly revealed and elegantly addressed, by learning the deconfounded object-relevant association for WSVOG. In the future, we will further explore the potential of causal intervention by designing more comprehensive intervention strategies on the variables in WSVOG. Besides, we will also incorporate external expert knowledge into our intervention \mbox{process to make up the lack of supervision signals.}

\balance
\bibliographystyle{IEEEtran}
%% argument is your BibTeX string definitions and bibliography database(s)
\bibliography{WSVOG}
\end{document}